\documentclass[lettersize,journal]{IEEEtran}
\usepackage{amsmath,amsfonts,amssymb}
\usepackage{array}
\usepackage{booktabs}
\usepackage{multirow}
\usepackage[caption=false,font=normalsize,labelfont=sf,textfont=sf]{subfig}
\usepackage{textcomp}
\usepackage{stfloats}
\usepackage{url}
\usepackage{graphicx}
\usepackage{cite}
\usepackage{xcolor}
\graphicspath{
  {figures/}
  {../Experiment1/test_outputs/pattern/}
  {../Experiment2/test_outputs/pattern/}
  {../Experiment3/test_outputs/pattern/}
}

\newcommand{\ee}{\mathrm{e}}
\newcommand{\jj}{\mathrm{j}}
\newcommand{\vect}[1]{\mathbf{#1}}
\newcommand{\mat}[1]{\mathbf{#1}}
\makeatletter
\newcommand{\Ms}{\@ifnextchar_{\Ms@sub}{\vect{M}}}
\def\Ms@sub_#1{\vect{M}_{#1}}
\makeatother

\begin{document}

\title{Physics-Informed Electromagnetic Modeling of\\
Hardware-Impaired Reconfigurable Intelligent Surfaces}

\author{Qiaojian Han,~\IEEEmembership{Student Member,~IEEE},  Yueheng Li,~\IEEEmembership{Member,~IEEE}, Haixia Zhang,~\IEEEmembership{Senior Member,~IEEE} and Dongfeng Yuan,~\IEEEmembership{Senior Member,~IEEE}%
\thanks{Qiaojian Han(qjhan@mail.sdu.edu.cn) is with the School of Control Science and Engineering, Shandong University, Jinan, China; Yueheng Li(yueheng.li@sdu.edu.cn) and Haixia Zhang(haixia.zhang@sdu.edu.cn) are with the Institute of Intelligent Communication Technology, Shandong University, Jinan, China; and Dongfeng Yuan(dfyuan@sdu.edu.cn) is with the School of Qilu Transportation, Shandong University, Jinan, China.}%
\thanks{All authors are also with the Shandong Key Laboratory of Intelligent Communication and Sensing-Computing Integration, China.}%
\thanks{Corresponding author: Haixia Zhang.}}

\maketitle

\begin{abstract}
Hardware impairments in practical reconfigurable intelligent surfaces (RISs) may cause a significant difference between the actual electromagnetic reflection responses and those modeled based on ideal hardware assumptions, leading to deviations on channel modeling of RIS-assisted wireless systems. There has been approaches considering hardware impairments based on analytical formulations and statistical assumptions in modeling hardware-impaired RISs. However, it is difficult to identify which assumptions best represent a practical RIS. What is more,  these models cannot capture various complex and coupled hardware impairments beyond those predefined analytical formulations. To address this issue, we propose a physics-informed electromagnetic modeling method with an embedded electromagnetic reflection forward model, which can learn the electric field magnitude patterns from hardware-impaired RISs. The method employs an ideal-state pretraining and impairment-aware fine-tuning strategy to separately learn actual RIS reflection behavior from physical mechanisms and data. Simulation  results show that the proposed model achieves  nice prediction with higher accuracy and faster convergence than those baselines.  In addition, it maintains high accuracy with a small amount of training data and achieves millisecond-level inference. Besides,  the predicted electric field magnitude pattern can be adopted to calculate the received power and the achievable rate, enabling accurate user rate prediction across representative hardware-impaired RIS-assisted communication scenarios.
\end{abstract}


\section{Introduction}
\IEEEPARstart{R}{econfigurable} 
intelligent surface (RIS) \cite{raeisi2024plugin,han2025physics,wang2024wideband} is of the abilities to control the reflection paths of electromagnetic waves to reconfigure wireless communication links, thus has been widely studied in millimeter-wave communications to help enhance their coverage areas.
 In most current studies, the electromagnetic reflection response of RISs is modeled as ideal phase-shift arrays \cite{delhougne2024generalized,jiang2025hybrid,chen2025ris}. However, with practical hardware impairments, the electromagnetic reflection response of RISs can be very different from that modeled under ideal RIS assumptions \cite{peng2024beamforming,boulogeorgos2020hardware}.
  Recent studies have reported an order-of-magnitude degradation in localization performance at high SNRs \cite{ozturk2022impact} and an overestimation of transmission  throughput by about 20\% \cite{chu2022ris} caused by ignoring hardware impairments of RIS. 

Existing studies have proposed various analytical and statistical models for RIS hardware impairments.
 Phase deviation and amplitude attenuation caused by manufacturing uncertainties are modeled using random phase distributions \cite{wang2023ris,li2023active} and random amplitude attenuation factors \cite{gomes2022tensor}, respectively.
  Further, S. Abeywickrama et al. proposed an amplitude-phase coupling model based on the circuit characteristics of RIS elements to characterize the coupling relationship between amplitude and phase deviations\cite{abeywickrama2020intelligent}.
   D. Wijekoon et al. developed mutual coupling models to characterize electromagnetic interactions among RIS elements \cite{wijekoon2023beamforming,wijekoon2024electromagnetic}.
    Beyond these hardware deviations, element stuck-at failure resulting from device degradation has also been considered and modeled in \cite{wang2023aging,yang2023phase}.
     These studies all show that hardware impairments arised from different physical origins and mechanisms lead to different analytical formulations.
      However, it is difficult to figure out which formulation best represents a practical  impairment.
        The most important is that those predefined analytical formulations cannot capture all kinds of complex and coupled hardware impairments, particularly those caused by the hardware-specific behavior of individual RIS implementations\cite{yang2023phase,cao2023insufficient}.
         This has motivated a new research topic on learning hardware impairments from measurable electromagnetic response dataset, such as the far electric field\cite{comite2018planar}.
          Purely data-driven learning may rely on a large amount of training data and is inefficient.
           To address this issue, combining known electromagnetic mechanisms with data-driven learning utilizing physics-informed neural networks (PINNs) \cite{raissi2019physics,bidabadi2025physically,tufa2026adaptive} can be  a promising approach.

Conventional PINNs adopt both physics and data loss functions in learning responses based on physical mechanisms and data.
 Sometimes, the two loss functions produce conflicting gradients during the training process, which lower the learning speed and accuracy of hardware impairment modeling\cite{wang2026Gradient}.
   Such conflicts can become intenser \cite{Zou2024Correcting} if there is a mismatch between ideal RIS reflection mechanisms and measured data.
    To address this issue, we propose a two-stage learning strategy which adopts an ideal-state pretraining and impairment-aware fine-tuning to separately learn actual RIS reflection behavior from both the physical mechanisms and the measured data.
     In RIS electromagnetic modeling, far electric field magnitude patterns across numerous phase configurations with varying electromagnetic boundary conditions should be predicted.
      Conventional PINNs are typically trained under fixed boundary conditions \cite{Wang2025Transfer}, and obviously cannot be adopted directly. 
      Parametric PINNs \cite{cho2024parameterized} or physics-informed neural operators \cite{li2024physics} can accommodate varying boundary conditions, thus fit for modeling RIS electromagnetic behavior well.
       But both of them, require training the electromagnetic field distribution throughout the whole three computational domain \cite{hu2025deepomamba, karumuri2026physics}.
       The enormous number of possible RIS phase configurations and the large spatial coverage in far-field prediction make such training impractical.
        To solve this problem, we first embed an electromagnetic forward model to learn the RIS magnetic current from different phase configurations. 
        Then we adopt a fixed Green function operator to map the learned magnetic current to the far electric field magnitude pattern\cite{sha2024greens,ERGUL20089898}.
         In this way, RIS electromagnetic responses can be learned and full-domain field reconstruction can be avoided.
        The main contributions of this work are summarized as follows:
\raggedbottom
\begin{itemize}
  \item A two-stage learning strategy comprising ideal-state pretraining and impairment-aware fine-tuning is estibalished to separately learn ideal RIS reflection behavior from physical mechanisms and hardware-specific deviations from data. Compared to the ablation variants, the proposed strategy achieves higher prediction accuracy and is of faster convergence.
  \item An electromagnetic reflection forward model is proposed and embedded to enable far electric field magnitude prediction across different RIS phase configurations without full-domain field reconstruction.
  \item A dataset construction method is developed to incorporate a broad range of analytical RIS hardware impairment models from existing studies and combine them. The resulting datasets are used to evaluate the proposed method in terms of sample efficiency, generalization, and inference latency. 
  \item Finally, a method calculating achievable rates from the predicted electric field magnitude patterns  is provided, which shows the extended applications of the predicted response. The achievable rates are calculated across four representative RIS-assisted communication scenarios.
\end{itemize}

\section{Electromagnetic Modeling and Neural Network}
This section describes the electromagnetic mechanism of RIS and introduces the neural representation of RIS electromagnetic responses. It then presents the physics-informed electromagnetic network structure, and establishes the calculation of achievable rate from the predicted far electric field magnitude.
\subsection{Electromagnetic Mechanism of the RIS}
The RIS electromagnetic reflection response  is derived by using Maxwell’s boundary conditions and the Green function. And the calculation of the equivalent magnetic current and the far electric field from the RIS phase configuration is also detailed in this part.

We consider a planar RIS with $M$ rows and $N$ columns. The element index set is $\{(i,j)\mid i=1,\ldots,M,\; j=1,\ldots,N\}$, and the total number of elements is $L=MN$. Let $\phi_{ij}$ denote the phase applied to element $(i,j)$, the RIS phase configuration is represented by
\begin{equation}
  \boldsymbol{\Phi}=[\phi_{ij}].
  \label{eq:bit_to_phase}
\end{equation}

We place the origin at the geometric center of the RIS. The $x$-axis follows the column index $j$, the $y$-axis follows the row index $i$.
 Under the plane rectangular coordinate system, the center coordinate of element $(i,j)$ is
\begin{equation}
  \begin{aligned}
  x_{ij}&=\left(j-\frac{N}{2}-\frac{1}{2}\right)d,\\
  y_{ij}&=\left(i-\frac{M}{2}-\frac{1}{2}\right)d,\\
  \vect{r}_{ij}&=[x_{ij},y_{ij},0]^T ,
  \end{aligned}
  \label{eq:coords}
\end{equation}
where $d$ is the side length of each RIS element  and $\vect{r}_{ij}$ is the position vector from the RIS coordinate origin to the center of the $(i,j)$-th element.

We extend the coordinate system to three dimensions by introducing the $z$-axis normal to the RIS plane. Let $(\theta_{\mathrm{in}},\varphi_{\mathrm{in}})$ denote the incidence direction, where $\theta_{\mathrm{in}}$ and $\varphi_{\mathrm{in}}$ are the polar angle measured from the positive $z$-axis and the azimuth angle in the $x$--$y$ plane, respectively. The incident unit wave vector direction is
\begin{equation}
  \hat{\vect{k}}_{\mathrm{in}}=
  [\sin\theta_{\mathrm{in}}\cos\varphi_{\mathrm{in}},
   \sin\theta_{\mathrm{in}}\sin\varphi_{\mathrm{in}},
   \cos\theta_{\mathrm{in}}]^{T}.
  \label{eq:khat}
\end{equation}
For carrier frequency $f$, the free-space wavelength and wavenumber are
\begin{equation}
  \lambda_0=\frac{c_0}{f},
  \qquad
  k_0=\frac{2\pi}{\lambda_0},
  \label{eq:wavenumber}
\end{equation}
where $c_0$ is the speed of light in free space and $k_0$ is the free-space wavenumber. The incident vector electric field at the $(i,j)$-th RIS element is
\begin{equation}
  \vect{E}_{\mathrm{inc},ij}
  =
  E_{\mathrm{in}}\vect{E}_0
  \ee^{\jj k_0\hat{\vect{k}}_{\mathrm{in}}^{T}\vect{r}_{ij}} ,
  \label{eq:einc}
\end{equation}
where $E_{\mathrm{in}}$ is the incident field magnitude, and $\vect{E}_0$ is the unit polarization vector used in the program, satisfying $\vect{E}_0^{T}\hat{\vect{k}}_{\mathrm{in}}=0$ and $\|\vect{E}_0\|_2=1$.
The reflection coefficient of the $(i,j)$-th RIS element is
\begin{equation}
  \Gamma_{ij}=\rho\ee^{\jj\phi_{ij}},
  \label{eq:gamma_ref}
\end{equation}
where $\rho$ is the reflection magnitude. Therefore, the reflected electric field can be written as
\begin{equation}
  \vect{E}_{\mathrm{ref},ij}
  =
  \Gamma_{ij}\vect{E}_{\mathrm{inc},ij}.
  \label{eq:eref}
\end{equation}
With the $z$-axis defined as the RIS normal direction, the unit surface normal is $\hat{\vect{n}}=\hat{\vect{z}}$. According to the electromagnetic boundary conditions derived from Maxwell's equations\cite{sha2024greens,ERGUL20089898}, the equivalent magnetic current of RIS is given by
\begin{equation}
  \Ms_{\mathrm{ideal},ij}
  =
  -\hat{\vect{n}}\times \vect{E}_{\mathrm{ref},ij}
  =
  [E_{\mathrm{ref},y,ij},-E_{\mathrm{ref},x,ij},0]^T.
  \label{eq:ms_bc}
\end{equation}
Here, the subscripts $x$ and $y$ denote the tangential field components along the coordinate axes in the RIS plane.

After obtaining the equivalent magnetic current, the next step is to map it to the far electric field. To describe the observation directions in the far field, a spherical coordinate system is adopted, parameterized by the polar angle $\theta$ and the azimuth angle $\varphi$.  The far-field angular grid is
\begin{equation}
  \begin{aligned}
  \theta_m
  &=
  \frac{(m-1)\pi}{2(N_{\theta}-1)},\\
  \varphi_n
  &=
  \frac{2\pi(n-1)}{N_{\varphi}-1},
  \end{aligned}
  \label{eq:angle_grid}
\end{equation}
for $m=1,\ldots,N_{\theta}$ and $n=1,\ldots,N_{\varphi}$, where $N_{\theta}$ and $N_{\varphi}$ are the numbers of polar and azimuth samples. The $(m,n)$-th discrete unit observation direction, namely the $(\theta_m,\varphi_n)$ direction, is
\begin{equation}
  \hat{\vect{r}}_{m,n}
  =
  \begin{bmatrix}
  \sin\theta_m\cos\varphi_n\\
  \sin\theta_m\sin\varphi_n\\
  \cos\theta_m
  \end{bmatrix}.
  \label{eq:obs_unit}
\end{equation}
At observation radius $R_{\mathrm{obs}}$, the observation point is
\begin{equation}
  \vect{r}_{m,n}
  =
  R_{\mathrm{obs}}\hat{\vect{r}}_{m,n}.
  \label{eq:obs_point}
\end{equation}
The vector from the center of the $(i,j)$-th RIS element to the observation point is
\begin{equation}
  \vect{R}_{ij}^{(m,n)}
  =
  \vect{r}_{m,n}-\vect{r}_{ij}.
  \label{eq:Rij_vec}
\end{equation}
The distance and the unit vector pointing from the element center to the observation point are given by
\begin{equation}
  R_{ij}^{(m,n)}
  =
  \|\vect{R}_{ij}^{(m,n)}\|_2,
  \qquad
  \hat{\vect{R}}_{ij}^{(m,n)}
  =
  \frac{\vect{R}_{ij}^{(m,n)}}{R_{ij}^{(m,n)}} .
  \label{eq:Rij_distance}
\end{equation}
In free space, the far electric field radiated by an equivalent magnetic current can be evaluated using the Green function derived from Maxwell's equations \cite{sha2024greens,ERGUL20089898}. The electric field vector in the $(m,n)$-th observation direction is given by
\begin{equation}
  \begin{aligned}
  \vect{E}_{m,n}
  &=
  \jj k_0
  \sum_{i=1}^{M}\sum_{j=1}^{N}
  \Big[
  \big(
  \hat{\vect{R}}_{ij}^{(m,n)}
  \times \Ms_{\mathrm{ideal},ij}
  \big)\\
  &\quad
  \times
  \hat{\vect{R}}_{ij}^{(m,n)}
  \Big]
  \frac{
  \ee^{-\jj k_0 R_{ij}^{(m,n)}}
  }{
  4\pi R_{ij}^{(m,n)}
  }d^2 ,
  \end{aligned}
  \label{eq:green}
\end{equation}
where $d^2$ is the element-area factor used by the discretized radiation integral. The magnitude in the $(m,n)$-th direction is
\begin{equation}
  E[m,n]
  =
  \|\vect{E}_{m,n}\|_2.
  \label{eq:discrete_magnitude}
\end{equation}
The electric field magnitude samples over the upper reflection half-space are then arranged on the angular grid to form the complete $\theta\in[0,\pi/2]$, $\varphi\in[0,2\pi]$ pattern:
\begin{equation}
  \mat{E}
  =
  [E[m,n]]\in\mathbb{R}^{N_{\theta}\times N_{\varphi}}.
  \label{eq:discrete_pattern}
\end{equation}
Thus, $\mat{E}$ denotes the electric field magnitude pattern derived from the above formulations by arranging all directional samples $E[m,n]$ on the angular grid spanning $\theta\in[0,\pi/2]$ and $\varphi\in[0,2\pi]$.

\subsection{Neural Representation of RIS Electromagnetic Responses}
Next, we develop a neural representation of RIS electromagnetic responses through supervised learning, using RIS phase configurations as inputs and the far electric field magnitude patterns as targets.

For a given RIS under fixed incident electric field conditions, the geometric, excitation, observation, and hardware parameters are also fixed. We collect them into a parameter set $\boldsymbol{\eta}$:
\begin{equation}
  \boldsymbol{\eta}
  =
  \{M,N,d,f,E_{\mathrm{in}},\rho,R_{\mathrm{obs}},
  \theta_{\mathrm{in}},\varphi_{\mathrm{in}},\boldsymbol{\xi}\},
  \label{eq:fixed_instance}
\end{equation}
where $d$ is the side length of each RIS element, $f$ is the carrier frequency, $E_{\mathrm{in}}$ is the incident field magnitude, $\rho$ is the reflection magnitude, $R_{\mathrm{obs}}$ is the observation radius, $(\theta_{\mathrm{in}},\varphi_{\mathrm{in}})$ is the incidence direction, and $\boldsymbol{\xi}$ denotes the set of hardware impairment parameters given by
\begin{equation}
  \begin{aligned}
  \boldsymbol{\xi}=\bigl(
  &\{\Delta\phi_{ij},z_{ij},\alpha_{ij},
  \chi_{ij},\phi_{ij}^{\mathrm{fix}}\}_{i=1,j=1}^{M,N},\\
  &\beta_{\min},a_{\beta},\phi_0,\mat{S}
  \bigr),
  \end{aligned}
  \label{eq:hardware_impairment_descriptor}
\end{equation}
where $\Delta\phi_{ij}$ is the element phase deviation, $\alpha_{ij}$ and $z_{ij}$ are the attenuation coefficient and its mask, $\chi_{ij}$ and $\phi_{ij}^{\mathrm{fix}}$ are the element failure mask and stuck-at phase state, $\beta_{\min}$, $a_{\beta}$, and $\phi_0$ are the amplitude-phase coupling parameters, and $\mat{S}$ is the mutual coupling scattering matrix. 

For the RIS specified by $\boldsymbol{\eta}$, the hardware impairment dataset consists of pairs of phase configurations and far electric field magnitude responses:

\begin{equation}
  \mathcal{D}_{\eta}
  =
  \{(\boldsymbol{\Phi}^{(q)},\mat{E}_{\eta}^{(q)})\}_{q=1}^{Q},
  \label{eq:hardware_dataset}
\end{equation}
where $Q$ is the number of samples, $\boldsymbol{\Phi}^{(q)}$ is the phase configuration of the $q$-th sample, and $\mat{E}_{\eta}^{(q)}$ is the far electric field magnitude obtained through simulation or measurement under the parameters $\boldsymbol{\eta}$. Based on this dataset, we define the supervised learning task as learning the mapping:
\begin{equation}
  \mathcal{F}_{\eta}:\boldsymbol{\Phi}\mapsto\mat{E}_{\eta}.
  \label{eq:learning_task_total}
\end{equation}

\begin{figure*}[!t]
\centering
\includegraphics[width=1\textwidth]{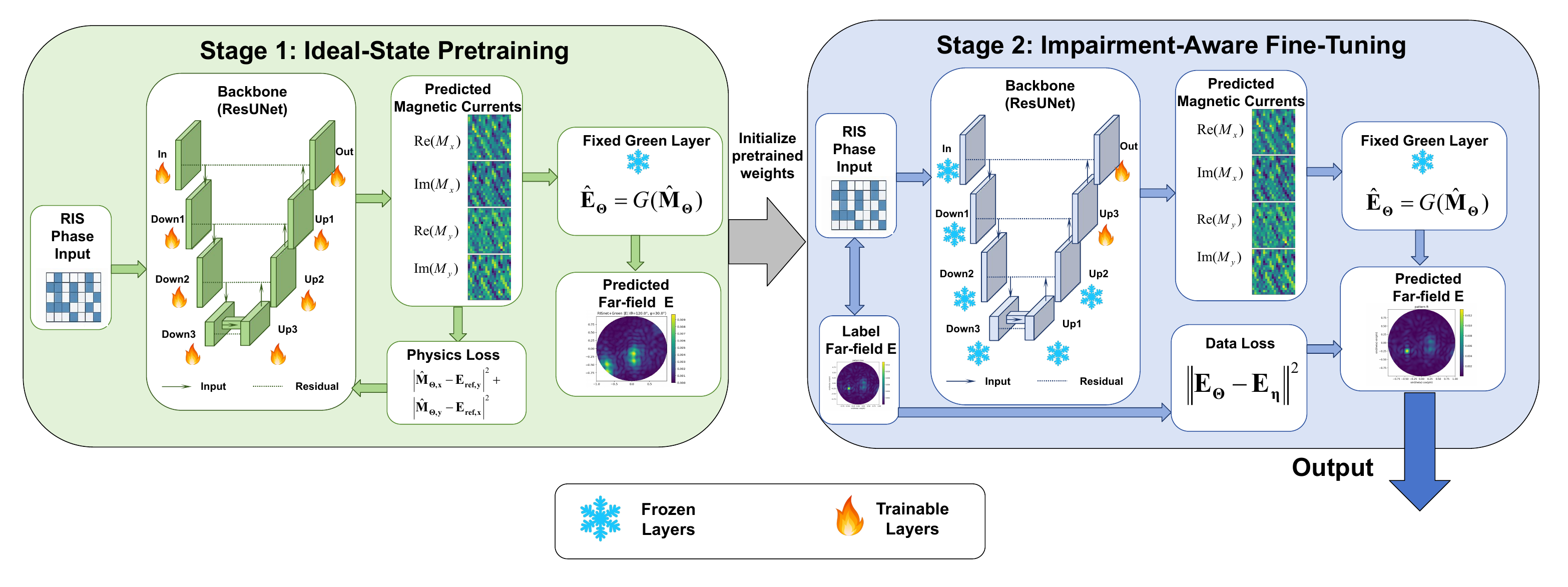}
\caption{The network structure and training protocol of the proposed model. The phase map is passed through the residual U-Net magnetic current predictor $g_{\boldsymbol{\Theta}}$, which predicts the complex tangential magnetic current. The predicted complex magnetic current is then propagated by the fixed Green function operator to obtain the far electric field magnitude pattern. During ideal-state pretraining, all parameters of $g_{\boldsymbol{\Theta}}$ are trainable and the loss is $\mathcal{L}_{\mathrm{phys}}$. During impairment-aware fine-tuning, most pretrained parameters are kept fixed and only a small trainable subset is updated using the data loss $\mathcal{L}_{\mathrm{data}}$.}
\label{fig:network_architecture}
\end{figure*}

\subsection{Physics-Informed Electromagnetic Network Structure}
After introducing the neural representation of RIS electromagnetic responses, we now focus on constructing a neural network architecture to solve the supervised learning problem. 
In the following,  the overall physics-informed structure of the proposed method, including the embedded electromagnetic forward model and the two-stage learning strategy, is presented in detail.

Let $\boldsymbol{\Theta}$ denote the trainable parameters of the magnetic current predictor, the end-to-end learned map can be written as
\begin{equation}
  \widehat{\mat{E}}_{\boldsymbol{\Theta}}
  =
  \widehat{\mathcal{F}}_{\boldsymbol{\Theta}}(\boldsymbol{\Phi}).
  \label{eq:nn_forward_total}
\end{equation}
To simplify the complex learning task in \eqref{eq:learning_task_total}, we use the magnetic current predictor $g_{\boldsymbol{\Theta}}$ to map the RIS phase configuration to the equivalent magnetic current, expressed as:
\begin{equation}
  \widehat{\mat{M}}_{\boldsymbol{\Theta}}
  = g_{\boldsymbol{\Theta}}(\boldsymbol{\Phi})
  = \left\{\widehat{\Ms}_{\boldsymbol{\Theta},ij}\right\}_{i=1,j=1}^{M,N},
  \label{eq:nn_ms_output}
\end{equation}
where $\widehat{\mat{M}}_{\boldsymbol{\Theta}}$ denotes the predicted equivalent magnetic currents of all RIS elements. At the $(i,j)$-th element, the magnetic current $\widehat{M}_{\boldsymbol{\Theta},ij}$ is:
\begin{equation}
  \widehat{\Ms}_{\boldsymbol{\Theta},ij}
  = [\widehat{M}_{\boldsymbol{\Theta},x,ij},
  \widehat{M}_{\boldsymbol{\Theta},y,ij},0]^T
  \in\mathbb{C}^{3},
  \label{eq:nn_complex_current}
\end{equation}
where $\widehat{M}_{\boldsymbol{\Theta},x,ij}$ and $\widehat{M}_{\boldsymbol{\Theta},y,ij}$ are the complex tangential components along the $x$- and $y$-axes, respectively.
Following the Green function in \eqref{eq:green}, the predicted magnetic current is mapped to the far electric field by replacing $\Ms_{\mathrm{ideal},ij}$ with $\widehat{\Ms}_{\boldsymbol{\Theta},ij}$:
\begin{equation}
\begin{aligned}
\widehat{\vect{E}}_{\boldsymbol{\Theta},m,n}
&=
\jj k_0
\sum_{i=1}^{M}\sum_{j=1}^{N}
\Big[
\big(
\hat{\vect{R}}_{ij}^{(m,n)}
\times
\widehat{\Ms}_{\boldsymbol{\Theta},ij}
\big)
\times
\hat{\vect{R}}_{ij}^{(m,n)}
\Big] \\
&\quad \times
\frac{
\ee^{-\jj k_0 R_{ij}^{(m,n)}}
}{
4\pi R_{ij}^{(m,n)}
}d^2 .
\end{aligned}
\label{eq:nn_green_operator}
\end{equation}
The predicted far electric field magnitude at the $(m,n)$-th observation direction is
\begin{equation}
\widehat{E}[m,n]
=
\left\|\widehat{\vect{E}}_{\boldsymbol{\Theta},m,n}\right\|_2.
\label{eq:nn_discrete_magnitude}
\end{equation}
The predicted far electric field magnitude pattern is obtained by arranging these directional samples over the angular grid:
\begin{equation}
\widehat{\mat{E}}_{\boldsymbol{\Theta}}
=
\left[
\widehat{E}[m,n]
\right]_{m=1,n=1}^{N_{\theta},N_{\varphi}}
\in\mathbb{R}^{N_{\theta}\times N_{\varphi}}.
\label{eq:nn_magnitude_pattern}
\end{equation}
For compactness, this embedded electromagnetic forward model is denoted as
\begin{equation}
\widehat{\mat{E}}_{\boldsymbol{\Theta}}
=
G
\left(
\widehat{\mat{M}}_{\boldsymbol{\Theta}}
\right),
\label{eq:embedded_forward_compact}
\end{equation}
where $G(\cdot)$ denotes the fixed electromagnetic mapping based on the Green function from the predicted equivalent magnetic current to the far electric field magnitude pattern.

Combining \eqref{eq:nn_ms_output} and \eqref{eq:embedded_forward_compact}, the prediction in \eqref{eq:nn_forward_total} is expressed as
\begin{equation}
\widehat{\mat{E}}_{\boldsymbol{\Theta}}
=
G\left(g_{\boldsymbol{\Theta}}(\boldsymbol{\Phi})\right).
\label{eq:multi_imp_network}
\end{equation}

The above formulations establish the overall architecture of the proposed physics-informed electromagnetic network, as illustrated in Fig. 1. 
Within this architecture, we implement the magnetic current predictor $g_{\boldsymbol{\Theta}}$ using a residual U-Net.
It consists of an input residual double-convolution block, three downsampling blocks, three upsampling blocks, and skip connections between the encoder and decoder stages. The output layer provides the predicted magnetic current defined in \eqref{eq:nn_complex_current}. Each residual double-convolution block contains two $3\times3$ convolutions with Group Normalization \cite{wu2018group} and sigmoid linear unit (SiLU) activation \cite{ramachandran2017searching}. A residual shortcut \cite{he2016deep} is added before the final SiLU activation. 

Next, we describe how to train this network using a two-stage learning strategy comprising ideal-state pretraining and impairment-aware fine-tuning.
 First, ideal-state pretraining trains the predictor to reproduce the equivalent magnetic current of RIS specified by \eqref{eq:ms_bc}.
  Then, impairment-aware fine-tuning starts from the pretrained predictor and updates its unfrozen parameters using hardware-impaired far-field data.
   The two stages are described in detail below.
\subsubsection{Pretraining Objective}
During ideal-state pretraining, all parameters of the magnetic current predictor $g_{\boldsymbol{\Theta}}$ are optimized. The adopted ideal-state pretraining problem is
\begin{equation}
  \boldsymbol{\Theta}_{0}^{\star}
  =
  \arg\min_{\boldsymbol{\Theta}}
  \mathcal{L}_{\mathrm{pre}}(\boldsymbol{\Theta}).
  \label{eq:pretrain_objective}
\end{equation}
where $\boldsymbol{\Theta}_{0}^{\star}$ denotes the network parameters through ideal-state pretraining without hardware impairments.
Let $\mathcal{Q}$ denote a mini-batch of commanded phase configurations.
 The physics loss compares the predicted tangential magnetic current components with the tangential reflected electric field components according to the boundary condition in \eqref{eq:ms_bc}, and is defined as
\begin{equation}
  \begin{aligned}
  \mathcal{L}_{\mathrm{phys}}
  &=
  \frac{1}{|\mathcal{Q}|MN}
  \sum_{q\in\mathcal{Q}}
  \sum_{i=1}^{M}
  \sum_{j=1}^{N}
  \Big(
  |\widehat{M}_{\boldsymbol{\Theta},x,ij}^{(q)}
  -E_{\mathrm{ref},y,ij}^{(q)}|^2\\
  &\hspace{2.5em}
  +
  |\widehat{M}_{\boldsymbol{\Theta},y,ij}^{(q)}
  +E_{\mathrm{ref},x,ij}^{(q)}|^2
  \Big).
  \end{aligned}
  \label{eq:physics_loss}
\end{equation}
Here $E_{\mathrm{ref},x,ij}^{(q)}$ and $E_{\mathrm{ref},y,ij}^{(q)}$ are the two tangential components of the reflected electric field in \eqref{eq:eref}.
The pretraining objective is
\begin{equation}
  \mathcal{L}_{\mathrm{pre}}
  =
  \mathcal{L}_{\mathrm{phys}}.
  \label{eq:pretrain_total_loss}
\end{equation}
During pretraining, gradients are backpropagated through the magnetic current predictor:
\begin{equation}
  \nabla_{\boldsymbol{\Theta}}\mathcal{L}_{\mathrm{pre}}
  =
  \sum_{q\in\mathcal{Q}}
  \frac{\partial\mathcal{L}_{\mathrm{phys}}}
       {\partial\widehat{\mat{M}}_{\boldsymbol{\Theta}}^{(q)}}
  \frac{\partial\widehat{\mat{M}}_{\boldsymbol{\Theta}}^{(q)}}
       {\partial\boldsymbol{\Theta}} .
  \label{eq:pretrain_gradient}
\end{equation}

\subsubsection{Fine-tuning Objective}
The impairment-aware fine-tuning stage starts from the pretrained parameters $\boldsymbol{\Theta}_{0}^{\star}$ obtained in \eqref{eq:pretrain_objective} and updates the unfrozen parameters using hardware-impaired far-field data.
Let $\boldsymbol{\Theta}_{\mathrm{ft}}$ and $\boldsymbol{\Theta}_{\mathrm{fr}}$ denote the network parameters selected for fine-tuning and the frozen parameters, respectively. Their pretrained values are denoted by $\boldsymbol{\Theta}_{0,\mathrm{ft}}^{\star}$ and $\boldsymbol{\Theta}_{0,\mathrm{fr}}^{\star}$. During fine-tuning, only $\boldsymbol{\Theta}_{\mathrm{ft}}$ is optimized, starting from $\boldsymbol{\Theta}_{0,\mathrm{ft}}^{\star}$, while $\boldsymbol{\Theta}_{\mathrm{fr}}$ remains at its pretrained value. The fine-tuning problem is written as
\begin{equation}
  \begin{aligned}
  \boldsymbol{\Theta}_{\mathrm{ft},\boldsymbol{\eta}}^{\star}
  &=
  \arg\min_{\boldsymbol{\Theta}_{\mathrm{ft}}}
  \mathcal{L}_{\mathrm{ft}}
  \bigl((\boldsymbol{\Theta}_{\mathrm{ft}},\boldsymbol{\Theta}_{0,\mathrm{fr}}^{\star});\boldsymbol{\eta}\bigr),\\
  \boldsymbol{\Theta}_{\mathrm{ft}}^{(0)}
  &=\boldsymbol{\Theta}_{0,\mathrm{ft}}^{\star}.
  \end{aligned}
  \label{eq:ft_objective}
\end{equation}
where $\boldsymbol{\eta}$ is the RIS parameter set defined in \eqref{eq:fixed_instance}. The complete fine-tuned parameter set is
$\boldsymbol{\Theta}_{\boldsymbol{\eta}}^{\star}=(\boldsymbol{\Theta}_{\mathrm{ft},\boldsymbol{\eta}}^{\star},\boldsymbol{\Theta}_{0,\mathrm{fr}}^{\star})$.
The data loss measures the mean squared error between the predicted far electric field magnitude patterns and the ground-truth patterns in the hardware-impaired dataset, and is defined as
\begin{equation}
  \mathcal{L}_{\mathrm{data}}
  =
  \frac{1}{|\mathcal{Q}|}
  \sum_{q\in\mathcal{Q}}
  \left\|
  \widehat{\mat{E}}_{\boldsymbol{\Theta}}^{(q)}
  -
  \mat{E}_{\eta}^{(q)}
  \right\|_F^2 .
  \label{eq:hw_data_loss}
\end{equation}
The impairment-aware fine-tuning objective is 
\begin{equation}
  \mathcal{L}_{\mathrm{ft}}
  =
  \mathcal{L}_{\mathrm{data}}.
  \label{eq:ft_total_loss}
\end{equation}

\subsection{Achievable Rate Calculation from the Predicted Electric Field}
The predicted far electric field magnitude $\widehat{E}[m,n]$ is used to calculate the received power and achievable rate for a user at distance $R_{\mathrm{obs}}$ from the RIS origin in direction $(\theta_m,\varphi_n)$.
The predicted magnitude $\widehat{E}[m,n]$ defined in \eqref{eq:nn_discrete_magnitude} follows the peak-phasor convention. The  time-average power density \cite{soderi2015NearField} at the $(m,n)$-th observation point is
\begin{equation}
  S_{\mathrm{av}}[m,n]
  =
  \frac{\widehat{E}^{2}[m,n]}{2\eta_0},
  \label{eq:average_power_density}
\end{equation}
where $S_{\mathrm{av}}[m,n]$ is measured in $\mathrm{W/m^2}$ and $\eta_0=\sqrt{\mu_0/\varepsilon_0}\approx376.73~\Omega$ is the free-space wave impedance. 

For a receive antenna with effective aperture $A_{\mathrm{e}}$, the received power \cite{jiang2023PhysicsBased} is
\begin{equation}
  P_{\mathrm{r}}[m,n]
  =
  S_{\mathrm{av}}[m,n]A_{\mathrm{e}},
  \qquad
  A_{\mathrm{e}}
  =
  \frac{G_{\mathrm{r}}\lambda_0^2}{4\pi},
  \label{eq:effective_aperture}
\end{equation}
where $G_{\mathrm{r}}$ is the linear receive antenna gain in the arrival direction and $\lambda_0=c_0/f$ is the carrier wavelength, with the speed of light in free space $c_0$ and carrier frequency $f$. Combining \eqref{eq:average_power_density} and \eqref{eq:effective_aperture} gives
\begin{equation}
  P_{\mathrm{r}}[m,n]
  =
  \frac{\widehat{E}^{2}[m,n]G_{\mathrm{r}}\lambda_0^2}
       {8\pi\eta_0}.
  \label{eq:received_power_from_field}
\end{equation}
The receiver noise power over a bandwidth of $B$ is
\begin{equation}
  P_{\mathrm{n}}
  =
  kTBF,
  \qquad
  F=10^{\mathrm{NF}/10},
  \label{eq:receiver_noise_power}
\end{equation}
where $k=1.380649\times10^{-23}~\mathrm{J/K}$ is the Boltzmann constant, $T$ is the system noise temperature, $B$ is bandwidth, $F$ is the linear receiver noise factor, and $\mathrm{NF}$ is the noise figure in decibels.

Small-scale fading is represented by the normalized complex coefficient $h_{\mathrm{f}}[m,n]$ at the $(m,n)$-th user location, with $\mathbb{E}[|h_{\mathrm{f}}[m,n]|^2]=1$\cite{zhang2019Calibrated,hosseinzadeh2025Privacy}. For Rayleigh fading, $h_{\mathrm{f}}[m,n]\sim\mathcal{CN}(0,1)$. For Rician fading, the normalized coefficient is modeled as the superposition of a deterministic line-of-sight (LoS) component and a scattered component:
\begin{equation}
  h_{\mathrm{f}}[m,n]
  =\sqrt{\frac{K}{K+1}}\,h_{\mathrm{LoS}}[m,n]
   +\sqrt{\frac{1}{K+1}}\,h_{\mathrm{sc}}[m,n],
\end{equation}
where $|h_{\mathrm{LoS}}[m,n]|=1$, $h_{\mathrm{sc}}[m,n]\sim\mathcal{CN}(0,1)$, and the Rician $K$-factor, $K\geq0$, is the ratio of the deterministic LoS power to the mean scattered power. 

The fading power gain is $G_{\mathrm{f}}[m,n]=|h_{\mathrm{f}}[m,n]|^2$. Applying this gain to the received power in \eqref{eq:received_power_from_field}, the instantaneous signal-to-interference-plus-noise ratio (SINR) is
\begin{equation}
  \gamma[m,n]
  =
  \frac{P_{\mathrm{r}}[m,n]G_{\mathrm{f}}[m,n]}
       {P_{\mathrm{I}}+P_{\mathrm{n}}}
  =
  \frac{\widehat{E}^{2}[m,n]G_{\mathrm{r}}\lambda_0^2G_{\mathrm{f}}[m,n]}
       {8\pi\eta_0(P_{\mathrm{I}}+kTBF)}.
  \label{eq:sinr_from_field}
\end{equation}
where $P_{\mathrm{I}}$ denotes the total interference power at the receiver.

Finally, the achievable rate associated with the $(m,n)$-th user location follows from the Shannon formula:
\begin{equation}
  \begin{aligned}
  R[m,n]
  &=
  B\log_2\!\left(1+\gamma[m,n]\right)\\
  &=
  B\log_2\!\left(
  1+
  \frac{\widehat{E}^{2}[m,n]G_{\mathrm{r}}\lambda_0^2G_{\mathrm{f}}[m,n]}
       {8\pi\eta_0(P_{\mathrm{I}}+kTBF)}
  \right),
  \end{aligned}
  \label{eq:achievable_rate_from_field}
\end{equation}
where $R[m,n]$ is measured in bit/s. 

\section{Hardware Impairment Modeling and Dataset Construction}
Following the hardware impairment reviewed in the introduction, this section presents their analytical models and describes how they are estiblished and combined. Then, the models are used to construct datasets of RIS electromagnetic responses.

\subsection{Phase Deviation}
For a phase deviation dataset, the commanded phase is shifted by a phase deviation $\Delta\phi_{ij}$. Thus, the element phase is
\begin{equation}
 \phi^{\mathrm{pd}}_{ij}
 =
 \phi_{ij}+\Delta\phi_{ij}.
 \label{eq:phase_act}
\end{equation}
The reflection coefficient used in \eqref{eq:gamma_ref} becomes
\begin{equation}
 \Gamma^{\mathrm{pd}}_{ij}
 =
 \rho\ee^{\jj\phi^{\mathrm{pd}}_{ij}}.
 \label{eq:phase_gamma}
\end{equation}
The uniform phase deviation model\cite{wang2023ris} is
\begin{equation}
 \Delta\phi_{ij}\sim\mathcal{U}[-\Delta_{\max},\Delta_{\max}],
 \label{eq:phase_uniform}
\end{equation}
and the Von Mises model \cite{li2023active} is
\begin{equation}
 f_{\Delta\phi}(\delta)
 =
 \frac{\ee^{\kappa\cos\delta}}{2\pi I_0(\kappa)},
 \quad \delta\in[-\pi,\pi),
 \label{eq:phase_vonmises}
\end{equation}
where $\Delta_{\max}$ is the maximum uniform phase deviation, $\kappa$ is the Von Mises concentration parameter, and $I_0(\cdot)$ is the zeroth-order modified Bessel function of the first kind. Adopting the uniform phase deviation model in \cite{wang2023ris} and the Von Mises model in \cite{li2023active}, we set $\Delta_{\max}=90^{\circ}$ and $\kappa=5$, to construct the phase deviation datasets. For both phase deviation datasets, $\Gamma_{ij}$ in \eqref{eq:gamma_ref} is replaced by $\Gamma^{\mathrm{pd}}_{ij}$ in \eqref{eq:phase_gamma}, and the far electric field response is then calculated using \eqref{eq:eref}--\eqref{eq:discrete_pattern}. 

\subsection{Amplitude Attenuation}
The amplitude attenuation dataset assigns each element an impairment mask
\begin{equation}
 z_{ij}\sim\operatorname{Bernoulli}(R_B),
 \label{eq:amp_mask}
\end{equation}
and an amplitude coefficient
\begin{equation}
 \alpha_{ij}
 =
 \begin{cases}
 1, & z_{ij}=0,\\
 r_{ij},\; r_{ij}\sim\mathcal{U}[\alpha_{\min},\alpha_{\max}], & z_{ij}=1 ,
 \end{cases}
 \label{eq:amp_alpha}
\end{equation}
where $R_B$ is the impaired-cell probability, $z_{ij}=1$ marks an impaired element, and $r_{ij}$ is the attenuation draw for that element.
The attenuation factor modifies the reflection coefficient as
\begin{equation}
 \Gamma^{\mathrm{amp}}_{ij}
 =
 \alpha_{ij}\rho\ee^{\jj\phi_{ij}}.
 \label{eq:amp_gamma}
\end{equation}
Adopting the amplitude attenuation model in \cite{gomes2022tensor}, we set $R_B=0.3$ and $[\alpha_{\min},\alpha_{\max}]=[0,1]$ to construct the amplitude attenuation dataset. $\Gamma_{ij}$ in \eqref{eq:gamma_ref} is replaced by $\Gamma^{\mathrm{amp}}_{ij}$ in \eqref{eq:amp_gamma}, and the  far electric field response is then calculated using \eqref{eq:eref}--\eqref{eq:discrete_pattern}.

\subsection{Amplitude-Phase Coupling}
The amplitude-phase coupling dataset uses a phase-dependent reflection magnitude. The amplitude function is
\begin{equation}
  \beta(\phi_{ij})
  =
  (1-\beta_{\min})
  \left(\frac{\sin(\phi_{ij}-\phi_0)+1}{2}\right)^{a_{\beta}}
  +
  \beta_{\min}.
  \label{eq:pda_beta}
\end{equation}
The parameters are  $\beta_{\min}=0.2$, $a_{\beta}=1.6$, and $\phi_0=0.43\pi$, where $\beta_{\min}$ is the minimum reflection amplitude, $a_{\beta}$ controls the steepness, and $\phi_0$ is the phase offset. The reflection coefficient is
\begin{equation}
 \Gamma^{\mathrm{apc}}_{ij}
 =
 \rho\beta(\phi_{ij})\ee^{\jj\phi_{ij}}.
 \label{eq:pda_gamma}
\end{equation}
The far electric field magnitude pattern is then constructed by using $\Gamma^{\mathrm{apc}}_{ij}$ in \eqref{eq:eref} and continuing through \eqref{eq:ms_bc}--\eqref{eq:discrete_pattern}.

\subsection{Mutual Coupling}
For notational convenience in the mutual coupling formulation, the two-dimensional element index $(i,j)$ is mapped to a one-dimensional index $\ell=(i-1)N+j$, where $\ell\in\{1,\ldots,L\}$ and $L=MN$. Unlike the preceding local impairments, mutual coupling mixes the reflected responses of different elements. For each commanded phase configuration, define the diagonal reflection matrix as 
\begin{equation}
 \mat{\Upsilon}
 =
 \operatorname{diag}(\ee^{\jj\phi_1},\ldots,\ee^{\jj\phi_L})
 \label{eq:upsilon}
\end{equation}
Let $\mat{S}\in\mathbb{C}^{L\times L}$ denote the mutual coupling scattering matrix, which describes electromagnetic coupling among the RIS elements. The effective reflection response accounting for this coupling is represented by
\begin{equation}
 \mat{P}_{\mathrm{mc}}
 =
 (\mat{\Upsilon}^{-1}-\mat{S})^{-1}.
 \label{eq:mc_operator}
\end{equation}
Equivalently,
\begin{equation}
 \mat{P}_{\mathrm{mc}}
 =
 (\mat{I}-\mat{\Upsilon}\mat{S})^{-1}\mat{\Upsilon},
 \label{eq:mc_equiv}
\end{equation}
where $\mat{I}$ is the $L\times L$ identity matrix. 

During dataset construction, the coupled fields are computed by solving the linear system associated with \eqref{eq:mc_equiv}.
The mutual coupling scattering matrix $\mat{S}\in\mathbb{C}^{L\times L}$ used in this system is constructed from the embedded-pattern coupling matrix $\mat{B}_{\mathrm{ep}}$ through its eigen decomposition:
\begin{equation}
 \begin{aligned}
 \mat{B}_{\mathrm{ep}}&=\mat{U}\mat{\Lambda}\mat{U}^{H},\\
 \mat{S}&=\mat{U}\sqrt{\mat{I}-\mat{\Lambda}}\mat{U}^{T}.
 \end{aligned}
 \label{eq:s_matrix}
\end{equation}
where $\mat{U}$ contains the eigenvectors of $\mat{B}_{\mathrm{ep}}$, and $\mat{\Lambda}$ is the diagonal matrix of the corresponding eigenvalues. The superscripts $H$ and $T$ denote Hermitian transpose and ordinary transpose, respectively. The entries of $\mat{B}_{\mathrm{ep}}$ are given by
\begin{equation}
 [\mat{B}_{\mathrm{ep}}]_{\ell k}
 =
 \begin{cases}
 \displaystyle \pi\left(\frac{d}{\lambda_0}\right)^2,
 & \ell=k,\\[6pt]
 \displaystyle \frac{d}{\lambda_0}
 \frac{J_1\!\left(2\pi(d/\lambda_0)d_{\ell k}\right)}{d_{\ell k}},
 & \ell\ne k.
 \end{cases}
 \label{eq:b_matrix}
\end{equation}
where $d_{\ell k}$ is the distance between elements $\ell$ and $k$ measured in units of the element spacing $d$, and $J_1(\cdot)$ is the Bessel function of the first kind of order one. With the resulting $\mat{S}$ incorporated into $\mat{P}_{\mathrm{mc}}$, the coupled reflected tangential fields are
\begin{equation}
 \vect{E}_{x,\mathrm{ref}}
 =
 \rho\mat{P}_{\mathrm{mc}}\vect{E}_{x,\mathrm{inc}},
 \quad
 \vect{E}_{y,\mathrm{ref}}
 =
 \rho\mat{P}_{\mathrm{mc}}\vect{E}_{y,\mathrm{inc}},
 \label{eq:mc_ref_fields}
\end{equation}
where $\vect{E}_{x,\mathrm{inc}}$ and $\vect{E}_{y,\mathrm{inc}}$ are the vectorized incident field components in the same $\ell$ order. These coupled reflected field components replace the uncoupled tangential components in \eqref{eq:ms_bc} and the far electric field magnitude is then obtained through \eqref{eq:angle_grid}--\eqref{eq:discrete_pattern}.

\subsection{Element Stuck-At Failure}
We consider element stuck-at failures caused by open or short circuit faults \cite{wang2023aging,yang2023phase}.
A failed element is fixed at either $0$ or $\pi$, independent of the commanded phase.
Let $\chi_{ij}\in\{0,1\}$ denote the failure mask, where $\chi_{ij}=1$ indicates a failed element, and let $\phi_{ij}^{\mathrm{fix}}\in\{0,\pi\}$ denote its fixed phase state.
 The actual phase of the $(i,j)$-th element is given by
\begin{equation}
 \phi^{\mathrm{st}}_{ij}
 =
 (1-\chi_{ij})\phi_{ij}
 +
 \chi_{ij}\phi_{ij}^{\mathrm{fix}} .
 \label{eq:pin_failure}
\end{equation}
The reflection coefficient is 
\begin{equation}
 \Gamma^{\mathrm{st}}_{ij}
 =
 \rho\ee^{\jj\phi^{\mathrm{st}}_{ij}}.
 \label{eq:pin_gamma}
\end{equation}
The far electric field response is obtained by replacing $\Gamma_{ij}$ in \eqref{eq:gamma_ref} with $\Gamma^{\mathrm{st}}_{ij}$ in \eqref{eq:pin_gamma}, and then calculating the reflected field, equivalent magnetic current, and far electric field using \eqref{eq:eref}--\eqref{eq:discrete_pattern}.

\begin{figure}[!t]
\centering
\includegraphics[width=\columnwidth]{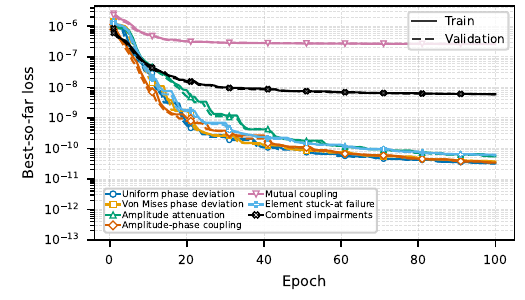}
\caption{Training and validation convergence for physics-only pretraining + impairment-aware fine-tuning across the seven hardware impairment datasets. Solid and dashed curves denote training and validation loss.}
\label{fig:exp1_training_curves}
\end{figure}

\begin{figure*}[!t]
\centering
\includegraphics[width=1\textwidth]{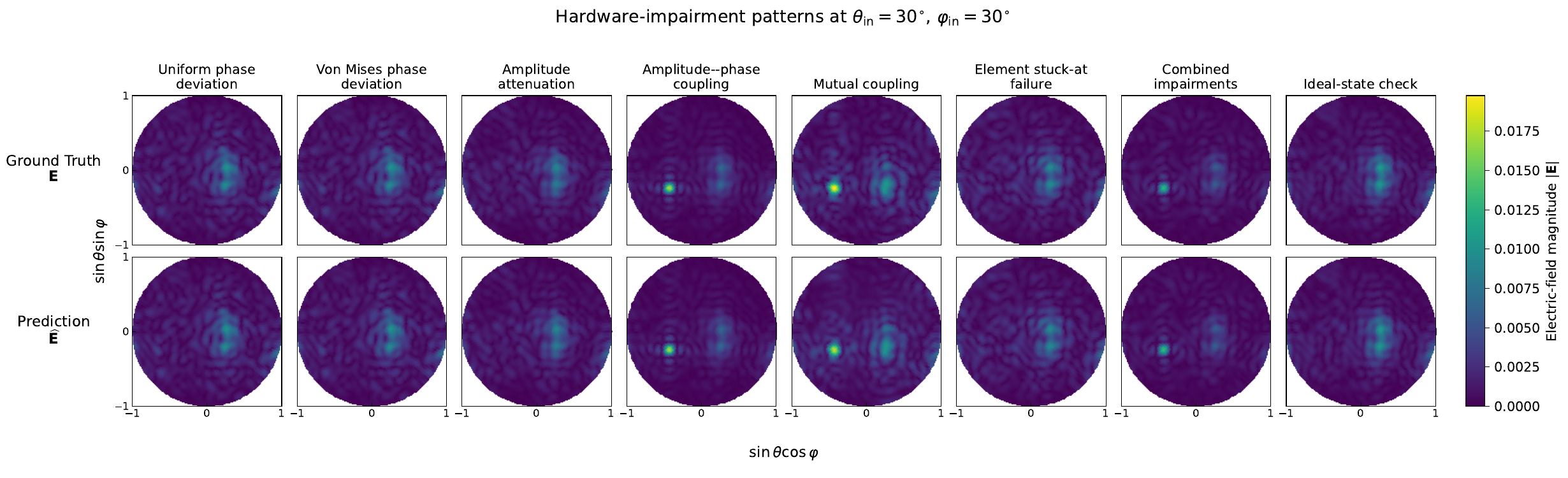}
\caption{Typical far electric field magnitude comparisons across the seven hardware impairment models and the ideal-state check. The upper row shows the ground-truth pattern $\mat{E}$, and the lower row shows the prediction $\widehat{\mat{E}}$.}
\label{fig:exp1_patterns}
\end{figure*}

\begin{table*}[!b]
\caption{Test Set Results Across Hardware Impairment Categories}
\label{tab:unified_results}
\centering
\begin{tabular}{lccc>{\bfseries\boldmath}c>{\bfseries\boldmath}c>{\bfseries\boldmath}c}
\toprule
\multirow{2}{*}{Dataset} &
\multicolumn{3}{c}{\shortstack{Ideal-state pretrained model\\without fine-tuning}} &
\multicolumn{3}{c}{\shortstack{\textbf{Physics-only pretraining +}\\\textbf{impairment-aware fine-tuning (proposed)}}} \\
\cmidrule(lr){2-4}\cmidrule(lr){5-7}
 & NMAE & NMSE & SSIM & NMAE & NMSE & SSIM \\
\midrule
Uniform phase deviation & $7.89{\times}10^{-2}$ & $1.05{\times}10^{-2}$ & 0.8093 & $1.70{\times}10^{-3}$ & $8.51{\times}10^{-6}$ & 0.9998 \\
Von Mises phase deviation & $7.89{\times}10^{-2}$ & $1.05{\times}10^{-2}$ & 0.8093 & $1.64{\times}10^{-3}$ & $6.90{\times}10^{-6}$ & 0.9999 \\
Amplitude attenuation & $7.04{\times}10^{-2}$ & $8.31{\times}10^{-3}$ & 0.8955 & $1.94{\times}10^{-3}$ & $1.19{\times}10^{-5}$ & 0.9998 \\
Amplitude-phase coupling & $4.36{\times}10^{-2}$ & $3.91{\times}10^{-3}$ & 0.1752 & $4.85{\times}10^{-4}$ & $9.91{\times}10^{-7}$ & 0.9999 \\
Mutual coupling & $3.97{\times}10^{-2}$ & $4.04{\times}10^{-3}$ & 0.2494 & $2.17{\times}10^{-2}$ & $8.13{\times}10^{-4}$ & 0.9250 \\
Element stuck-at failure & $7.72{\times}10^{-2}$ & $9.97{\times}10^{-3}$ & 0.8222 & $2.93{\times}10^{-3}$ & $2.88{\times}10^{-5}$ & 0.9995 \\
Combined impairments & $6.84{\times}10^{-2}$ & $8.24{\times}10^{-3}$ & 0.2167 & $5.46{\times}10^{-3}$ & $4.90{\times}10^{-5}$ & 0.9947 \\
\bottomrule
\end{tabular}
\end{table*}

\begin{table}[!b]
\caption{Ideal-State Pretraining Ablation}
\label{tab:ideal_pretraining_ablation}
\centering
\footnotesize
\setlength{\tabcolsep}{3pt}
\begin{tabular}{>{\raggedright\arraybackslash}p{0.49\columnwidth}cc}
\toprule
Variant & \shortstack{Best validation\\data loss} & \shortstack{Training\\time (s)} \\
\midrule
Backbone-only direct-field training & $8.66{\times}10^{-7}$ & 1122.6 \\
Physics-structured data-only training & $8.85{\times}10^{-7}$ & 8589.0 \\
Physics-structured data+physics training & $9.49{\times}10^{-12}$ & 8648.8 \\
Physics-only pretraining & $\mathbf{4.22{\times}10^{-12}}$ & $\mathbf{328.7}$ \\
\bottomrule
\end{tabular}
\end{table}

\subsection{Combined Impairment}
The combined impairment model first accounts for element stuck-at failures. For each commanded one-bit phase configuration, the actual element phase follows the vectorized form of \eqref{eq:pin_failure}:
\begin{equation}
 \phi_{\ell}^{\mathrm{st}}
 = (1-\chi_{\ell})\phi_{\ell}
 + \chi_{\ell}\phi_{\ell}^{\mathrm{fix}},
 \label{eq:combined_stuck_phase}
\end{equation}
where $\chi_{\ell}$ and $\phi_{\ell}^{\mathrm{fix}}$ are the vectorized failure mask and fixed phase state, respectively. The stuck-at rate is 10\%, with equal numbers of failed elements fixed at $0$ and $\pi$.

Phase deviations are then added:
\begin{equation}
 \phi_{\ell}^{\mathrm{all}}
 =
 \phi_{\ell}^{\mathrm{st}}+\Delta\phi_{\ell},
 \label{eq:combined_phase}
\end{equation}
where $\Delta\phi_{\ell}$ is drawn uniformly from $[-11.25^{\circ},11.25^{\circ}]$.

Amplitude attenuation and amplitude-phase coupling are incorporated into the local reflection response as
\begin{equation}
 v_{\ell}^{\mathrm{all}}
 =
 \alpha_{\ell}
 \beta(\phi_{\ell}^{\mathrm{all}})
 \ee^{\jj\phi_{\ell}^{\mathrm{all}}},
 \label{eq:combined_local_coeff}
\end{equation}
where $\alpha_{\ell}$ is drawn from $[0.95,1]$, and $\beta(\phi_{\ell}^{\mathrm{all}})$ is given by \eqref{eq:pda_beta} with $\beta_{\min}=0.2$, $a_{\beta}=1.6$, and $\phi_0=0.43\pi$. The local responses are assembled into
\begin{equation}
 \mat{\Upsilon}_{\mathrm{all}}
 =
 \operatorname{diag}
 \left(v_{1}^{\mathrm{all}},\ldots,v_{L}^{\mathrm{all}}\right).
 \label{eq:combined_upsilon}
\end{equation}
Finally, mutual coupling is incorporated by substituting $\mat{\Upsilon}_{\mathrm{all}}$ into \eqref{eq:mc_operator}, with $\mat{S}$ constructed using \eqref{eq:s_matrix} and \eqref{eq:b_matrix}. The coupled reflected fields are calculated using \eqref{eq:mc_ref_fields}, converted to the equivalent magnetic current through \eqref{eq:ms_bc}, and propagated to the far electric field response using \eqref{eq:green}--\eqref{eq:discrete_pattern}.

\subsection{Dataset Construction}
\label{sec:dataset_construction}
We construct a dataset for each impairment model presented in the preceding subsections: uniform phase deviation, Von Mises phase deviation, amplitude attenuation, amplitude-phase coupling, mutual coupling, element stuck-at failure, and combined impairment. Each dataset contains 5000 independently and randomly generated commanded one-bit RIS phase configurations, each paired with the corresponding far electric field magnitude pattern calculated using the associated impairment model. 
Each dataset is divided into three disjoint subsets containing 2000 training, 1000 validation, and 2000 test samples, with no commanded phase configuration shared between subsets.
\section{Simulation Results}
This section presents simulation results for the proposed method.
 Following the simulation setup and evaluation metrics, we assess prediction accuracy under different hardware impairments.
  Ablation studies examine the contributions of the network structure and the two-stage training strategy, with inference latency reported for each variant.
   Then, we evaluate sample efficiency with limited hardware-impaired training data and generalization to unseen incidence angles.
\subsection{Simulation Setup and Evaluation Metrics}
The simulations use an RIS with $M=N=20$, element spacing $d=5$~mm, carrier frequency $f=28$~GHz, incident field magnitude $E_{\mathrm{in}}=1$, reflection magnitude $\rho=1$, and observation radius $R_{\mathrm{obs}}=10$~m. With aperture width $D=(M-1)d=95$~mm, the Fraunhofer distance is $R_{\mathrm{FF}}=2D^2/\lambda_0\approx1.69$~m; hence, $R_{\mathrm{obs}}=10$~m lies in the far-field region. The far electric field magnitude is sampled over the upper reflection half-space with $\theta\in[0^{\circ},90^{\circ}]$ and the full azimuth range $\varphi\in[0^{\circ},360^{\circ}]$. With $N_{\theta}=90$ and $N_{\varphi}=180$, the nominal angular grid resolutions are approximately $1^{\circ}$ and $2^{\circ}$.
 Unless otherwise stated, the incidence angle is
$(\theta_{\mathrm{in}},\varphi_{\mathrm{in}})=(30^{\circ},30^{\circ})$.

The networks are trained using the Adam optimizer.
Ideal-state pretraining runs for 1000 epochs with a batch
size of 16 and an initial learning rate of $10^{-3}$.
Impairment-aware fine-tuning runs for 100 epochs with a
batch size of 64 and an initial learning rate of
$5\times10^{-3}$ for the unfrozen backbone parameters.
During fine-tuning, the validation loss is evaluated after
each epoch, and the learning rate is multiplied by $0.8$
when the validation loss fails to improve for more than
three consecutive epochs. All model training and inference were performed on a NVIDIA RTX A6000 GPU.

\begin{figure*}[!t]
\centering
\captionsetup[subfloat]{font=scriptsize}
\subfloat[Ideal-state pretraining data loss.]{\includegraphics[width=0.4\textwidth]{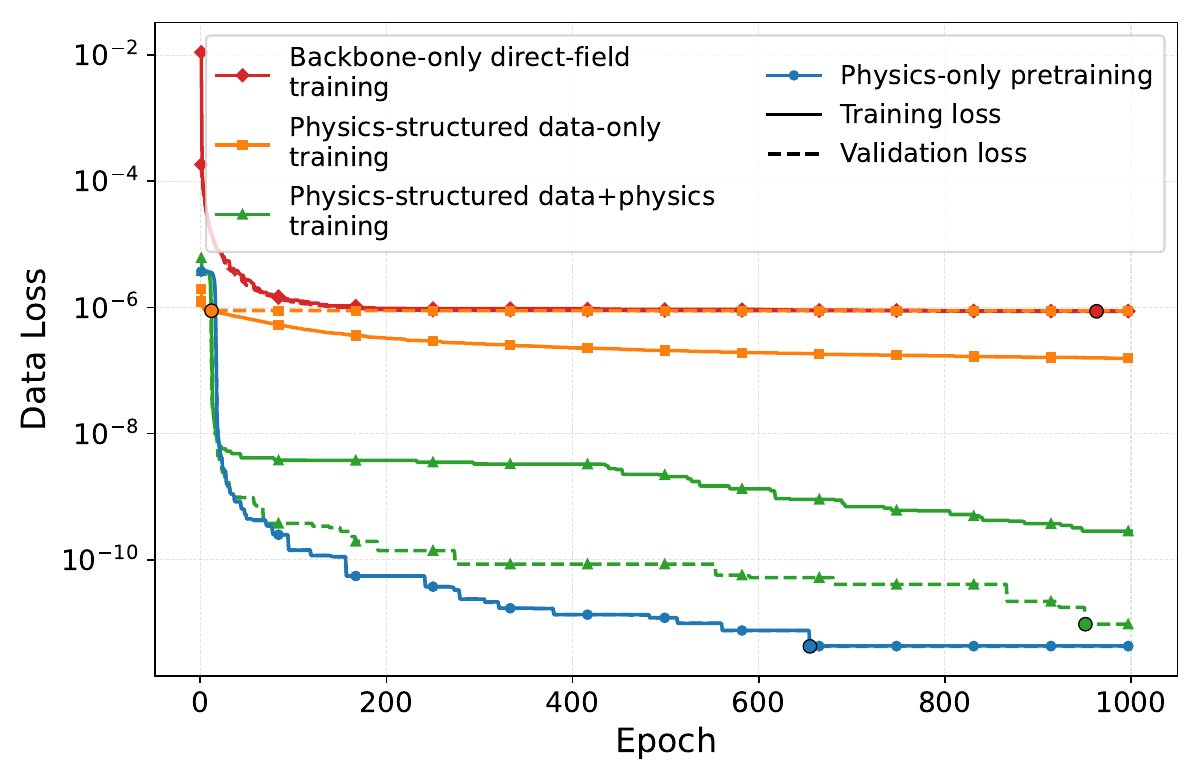}}
\hfill
\subfloat[Impairment-aware fine-tuning data loss.]{\includegraphics[width=0.4\textwidth]{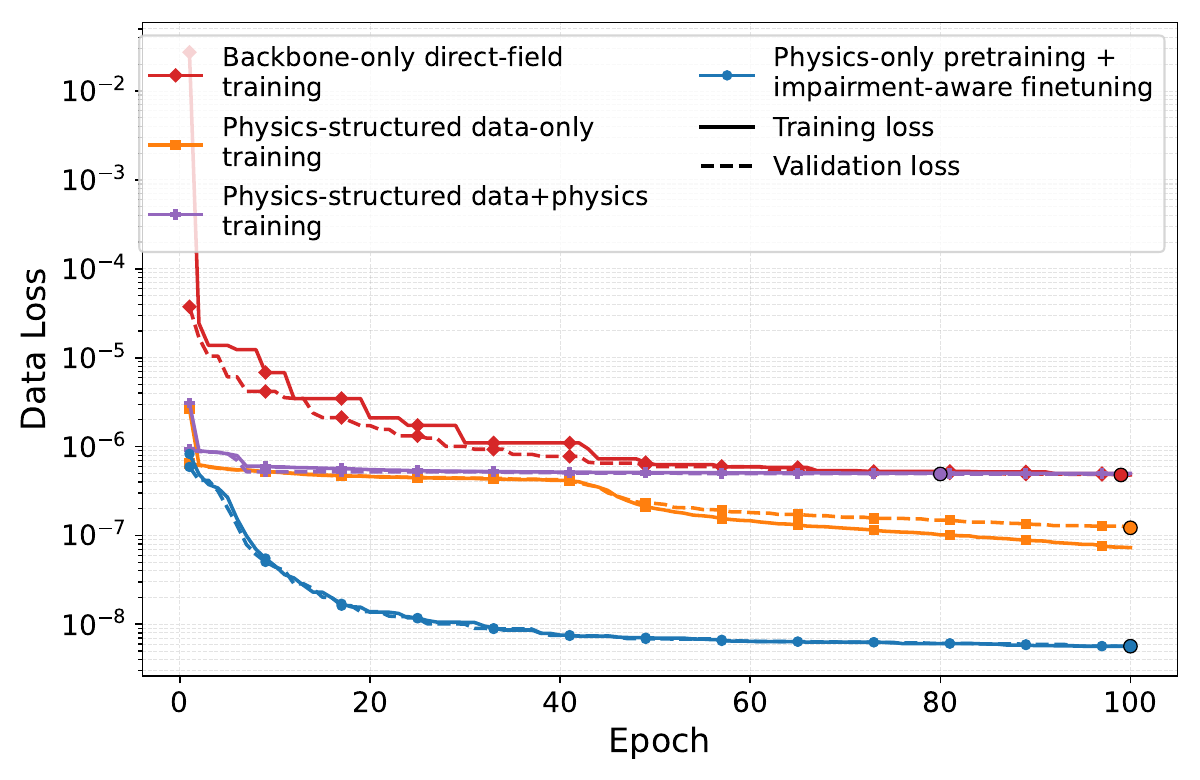}}
\caption{Two-stage best-so-far data loss convergence. Physics-only pretraining reaches the lowest validation data loss without using ideal-state pattern labels, while physics-only pretraining + impairment-aware fine-tuning reaches the lowest validation data loss during hardware-impaired adaptation.}
\label{fig:ideal_pretraining_ablation}
\end{figure*}

For a predicted far electric field magnitude pattern $\widehat{\mat{E}}_{\boldsymbol{\Theta}}$ and ground-truth pattern $\mat{E}_{\eta}$, let $E_{\eta}[m,n]$ denote the $(m,n)$-th entry of $\mat{E}_{\eta}$, corresponding to the predicted magnitude $\widehat{E}[m,n]$ defined in \eqref{eq:nn_discrete_magnitude}. The normalized mean absolute error (NMAE) is
\begin{equation}
  \operatorname{NMAE}
  =
  \frac{1}{N_{\theta}N_{\varphi}}
  \frac{
  \sum_{m=1}^{N_{\theta}}
  \sum_{n=1}^{N_{\varphi}}
  |\widehat{E}[m,n]-E_{\eta}[m,n]|
  }
  {\max_{m,n}|E_{\eta}[m,n]|},
  \label{eq:nmae_metric}
\end{equation}
The mean-square error (MSE) and normalized mean-square error (NMSE) are
\begin{equation}
  \operatorname{MSE}
  =
  \frac{1}{N_{\theta}N_{\varphi}}
  \sum_{m=1}^{N_{\theta}}
  \sum_{n=1}^{N_{\varphi}}
  |\widehat{E}[m,n]-E_{\eta}[m,n]|^2,
  \label{eq:mse_metric}
\end{equation}
\begin{equation}
  \operatorname{NMSE}
  =
  \frac{\operatorname{MSE}}
       {(\max_{m,n}|E_{\eta}[m,n]|)^2}.
  \label{eq:nmse_metric}
\end{equation}
The reported structural similarity index measure (SSIM) is
\begin{equation}
  \begin{aligned}
  \operatorname{SSIM}(\widehat{\mat{E}}_{\boldsymbol{\Theta}},\mat{E}_{\eta})
  &=
  \frac{(2\mu_{\widehat{E}}\mu_{E_{\eta}}+c_1)
        (2\sigma_{\widehat{E},E_{\eta}}+c_2)}
       {(\mu_{\widehat{E}}^2+\mu_{E_{\eta}}^2+c_1)
        (\sigma_{\widehat{E}}^2+\sigma_{E_{\eta}}^2+c_2)}.
  \end{aligned}
  \label{eq:ssim_formula}
\end{equation}
Here, $\mu_{\widehat{E}}$ and $\mu_{E_{\eta}}$ are the means of the  samples in $\widehat{\mat{E}}_{\boldsymbol{\Theta}}$ and $\mat{E}_{\eta}$. $\sigma_{\widehat{E}}^2$ and $\sigma_{E_{\eta}}^2$ are the variances of these samples in the respective patterns, $\sigma_{\widehat{E},E_{\eta}}$ is the covariance between the predicted and ground-truth magnitudes at matching angular grid points, and $c_1=10^{-4}$ and $c_2=9\times10^{-4}$ are stabilizing constants.

\subsection{Adaptability to Different Hardware Impairment Models}

This evaluation examines whether a unified physics-informed forward modeling structure can characterize multiple hardware impairments. 
The datasets constructed in Section~III are used to evaluate prediction accuracy under different hardware impairments. 
For each hardware impairment dataset, fine-tuning starts from the same pretrained parameters and uses the 2000 training samples. 

Figure~\ref{fig:exp1_training_curves} plots the training and validation loss over epochs. Across multiple hardware impairment datasets, both losses approach convergence within 100 epochs.
Table~\ref{tab:unified_results} reports their test set performance, demonstrating that the proposed method maintains high prediction accuracy across different types of hardware impairments, while ignoring these impairments leads to substantial performance degradation.
Figure~\ref{fig:exp1_patterns} shows comparisons between the predicted and ground-truth far electric field magnitude patterns for representative test samples from each hardware impairment dataset.

\begin{table*}[!t]
\caption{Hardware-Impaired Ablation and Per-Sample Inference Efficiency on the Combined Impairment Test Set}
\label{tab:baseline_results}
\centering
\begin{tabular}{lcccc}
\toprule
Variant & NMAE & NMSE & SSIM & Latency (ms) \\
\midrule
Backbone-only direct-field training & $4.08{\times}10^{-2}$ & $3.70{\times}10^{-3}$ & 0.097 & 0.21 \\
Physics-structured data-only training & $2.23{\times}10^{-2}$ & $9.99{\times}10^{-4}$ & 0.884 & 5.90 \\
Physics-structured data+physics training & $4.67{\times}10^{-2}$ & $4.33{\times}10^{-3}$ & 0.271 & 5.89 \\
Ideal-state pretrained model without fine-tuning & $6.84{\times}10^{-2}$ & $8.24{\times}10^{-3}$ & 0.217 & 5.89 \\
\textbf{Physics-only pretraining + impairment-aware fine-tuning (proposed)} & $\mathbf{5.46{\times}10^{-3}}$ & $\mathbf{4.90{\times}10^{-5}}$ & $\mathbf{0.995}$ & 5.89 \\
\bottomrule
\end{tabular}
\end{table*}

\begin{figure}[!t]
\centering
\includegraphics[width=0.42\textwidth]{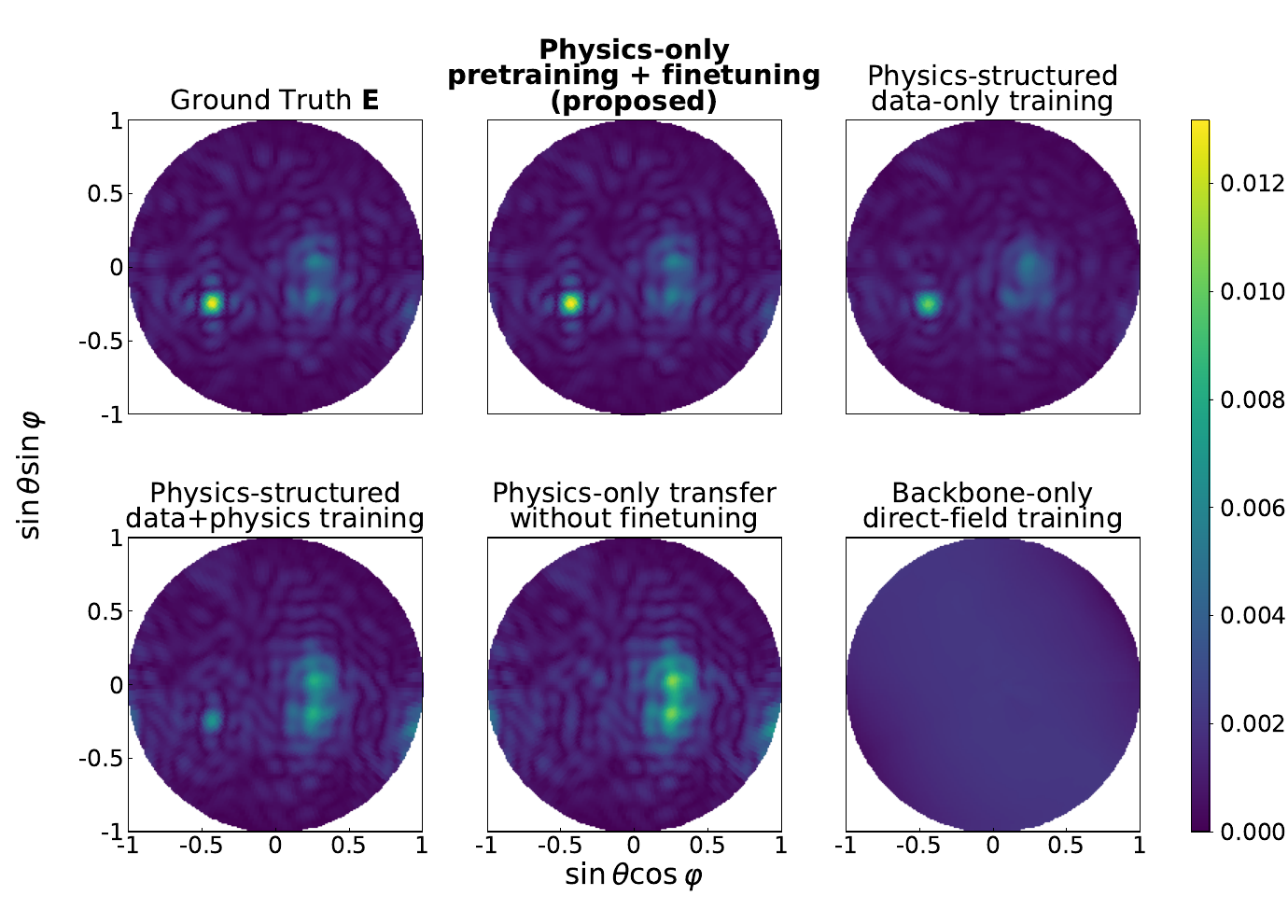}
\caption{Typical combined impairment sample for the hardware-impaired adaptation ablation. The ground-truth pattern $\mat{E}$ is placed next to the prediction from physics-only pretraining + impairment-aware fine-tuning, followed by the four competing variants. All panels use the same magnitude scale.}
\label{fig:ablation_pattern_comparison}
\end{figure}

\subsection{Two-Stage Ablation}
We conduct ablation studies for ideal-state pretraining and impairment-aware fine-tuning to examine how the training strategy and network structure affect prediction accuracy and convergence at each stage. The ideal-state pretraining ablation compares different network structures and training losses to investigate the effect of physics-informed pretraining. The hardware-impaired adaptation ablation evaluates the impact of the fine-tuning or training strategy on the prediction performance on the combined impairment dataset.
\subsubsection{Ideal-State Pretraining Ablation}
Four ideal-state training variants are compared:
(i) backbone-only direct-field training, which directly
predicts the far electric field magnitude pattern using
the data loss;
(ii) physics-structured data-only training, which includes
the embedded electromagnetic forward model and uses only
the data loss;
(iii) physics-structured data+physics training, which uses
the same architecture with both the data and physics losses;
and (iv) physics-only pretraining, which trains the magnetic
current predictor using only the physics loss.
Figure~\ref{fig:ideal_pretraining_ablation} compares the
best-so-far data loss during ideal-state pretraining and
the subsequent impairment-aware fine-tuning stage.

\begin{table*}[!t]
\caption{Sample Efficiency Results on the Combined Impairment Dataset}
\label{tab:sample_efficiency}
\centering
\renewcommand{\arraystretch}{0.80}
\begin{tabular}{cccc>{\bfseries\boldmath}c>{\bfseries\boldmath}c>{\bfseries\boldmath}c}
\toprule
\multirow{2}{*}{Train samples} &
\multicolumn{3}{c}{\shortstack{Physics-structured\\data-only training}} &
\multicolumn{3}{c}{\shortstack{\textbf{Physics-only pretraining +}\\\textbf{impairment-aware fine-tuning (proposed)}}} \\
\cmidrule(lr){2-4}\cmidrule(lr){5-7}
 & NMAE & NMSE & SSIM & NMAE & NMSE & SSIM \\
\midrule
100 & $5.02{\times}10^{-2}$ & $5.52{\times}10^{-3}$ & 0.131 & $4.34{\times}10^{-2}$ & $3.30{\times}10^{-3}$ & 0.653 \\
160 & $4.81{\times}10^{-2}$ & $5.23{\times}10^{-3}$ & 0.134 & $3.00{\times}10^{-2}$ & $1.60{\times}10^{-3}$ & 0.893 \\
200 & $4.74{\times}10^{-2}$ & $5.03{\times}10^{-3}$ & 0.168 & $1.47{\times}10^{-2}$ & $3.69{\times}10^{-4}$ & 0.964 \\
400 & $4.82{\times}10^{-2}$ & $5.22{\times}10^{-3}$ & 0.170 & $9.99{\times}10^{-3}$ & $1.90{\times}10^{-4}$ & 0.979 \\
600 & $4.86{\times}10^{-2}$ & $4.82{\times}10^{-3}$ & 0.371 & $7.45{\times}10^{-3}$ & $9.38{\times}10^{-5}$ & 0.990 \\
1000 & $4.59{\times}10^{-2}$ & $4.46{\times}10^{-3}$ & 0.312 & $6.50{\times}10^{-3}$ & $7.05{\times}10^{-5}$ & 0.992 \\
1600 & $2.85{\times}10^{-2}$ & $1.55{\times}10^{-3}$ & 0.817 & $5.56{\times}10^{-3}$ & $5.08{\times}10^{-5}$ & 0.994 \\
2000 & $2.23{\times}10^{-2}$ & $9.99{\times}10^{-4}$ & 0.884 & $5.46{\times}10^{-3}$ & $4.90{\times}10^{-5}$ & 0.995 \\
\bottomrule
\end{tabular}
\end{table*}

Table~\ref{tab:ideal_pretraining_ablation} shows that physics-only pretraining achieves the lowest loss and the shortest training time among all variants. It also performs better than physics-structured data+physics training, even without using ideal-state far-field data and labels.
 Therefore, we adopt physics-only pretraining for the proposed model.

\subsubsection{Hardware-Impaired Adaptation Ablation}
The second-stage comparison uses the fixed combined impairment dataset and its 2000-sample test split. The five variants are: (i) backbone-only direct-field training on hardware-impaired labels, which maps the commanded phase directly to the far-field pattern without the magnetic current and Green function structure; (ii) physics-structured data-only training from scratch; (iii) physics-structured data+physics training from scratch; (iv) the ideal-state pretrained model without fine-tuning; and (v) physics-only pretraining + impairment-aware fine-tuning. 

Table~\ref{tab:baseline_results} reports the prediction accuracy
of the five variants. Physics-only pretraining +
impairment-aware fine-tuning achieves the lowest NMAE and
NMSE and the highest SSIM.
Figure~\ref{fig:ablation_pattern_comparison} presents results
for a typical combined impairment test sample, showing that
the proposed method closely reproduces both the spatial
field structure and magnitude distribution of the ground truth.

Table~\ref{tab:baseline_results} also reports the average
per-sample inference latency under the protocol specified
in Section~IV-A. The proposed method requires 5.89~ms per
sample. Backbone-only direct-field training is faster because
it omits Green function propagation, whereas the
physics-structured variants include both magnetic current
prediction and fixed Green function propagation.

\subsection{Sample Efficiency}
To examine sample efficiency, we vary the number of hardware-impaired training samples. The sample efficiency evaluation uses the same 2000-sample training split of the combined impairment dataset as the training strategy comparison. Subsets containing 5\%, 8\%, 10\%, 20\%, 30\%, 50\%, 80\%, and 100\% of the training split are selected, corresponding to 100, 160, 200, 400, 600, 1000, 1600, and 2000 samples, while the validation and test sets remain fixed at 1000 and 2000 samples, respectively.
Physics-structured data-only training achieves higher prediction accuracy than the other baseline methods in Table~\ref{tab:baseline_results}. Therefore, we use it as the baseline for comparison with the proposed two-stage strategy in the sample efficiency evaluation.
 Table~\ref{tab:sample_efficiency} shows that physics-only pretraining + impairment-aware fine-tuning outperforms physics-structured data-only training at every ratio.

\begin{table}[!t]
\caption{Results at Unseen Incidence Angles}
\label{tab:angle_generalization_results}
\centering
\scriptsize
\setlength{\tabcolsep}{2.2pt}
\begin{tabular}{ccccc}
\toprule
\shortstack{Test angle\\$(\theta_{\mathrm{in}},\varphi_{\mathrm{in}})$} & Samples & NMAE & NMSE & SSIM \\
\midrule
$(15^{\circ},0^{\circ})$ & 100 & $2.75{\times}10^{-2}$ & $1.29{\times}10^{-3}$ & 0.892 \\
$(30^{\circ},60^{\circ})$ & 100 & $3.11{\times}10^{-2}$ & $1.64{\times}10^{-3}$ & 0.829 \\
$(30^{\circ},45^{\circ})$ & 100 & $2.98{\times}10^{-2}$ & $1.51{\times}10^{-3}$ & 0.837 \\
$(60^{\circ},180^{\circ})$ & 100 & $2.90{\times}10^{-2}$ & $1.44{\times}10^{-3}$ & 0.858 \\
$(45^{\circ},30^{\circ})$ & 100 & $3.18{\times}10^{-2}$ & $1.71{\times}10^{-3}$ & 0.824 \\
\bottomrule
\end{tabular}
\end{table}

\begin{figure*}[!b]
\centering
\includegraphics[width=0.8\textwidth]{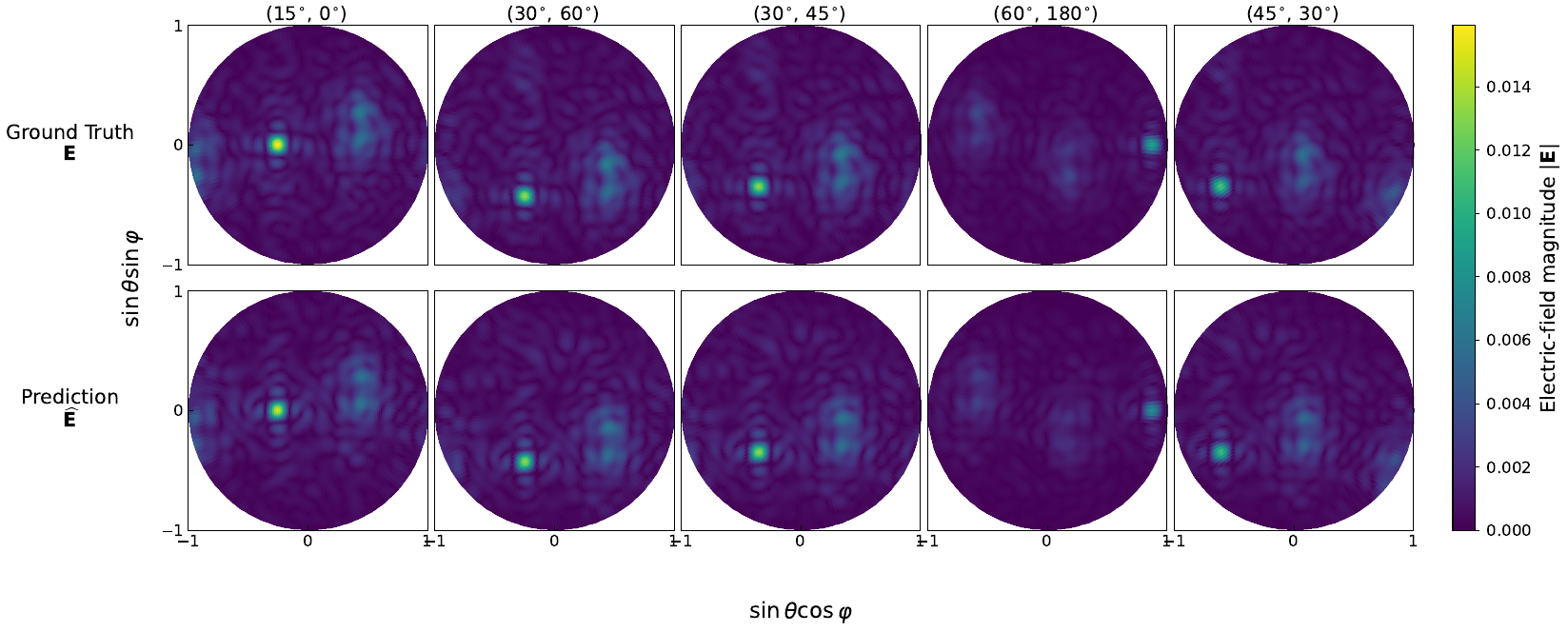}
\caption{Typical far electric field magnitude patterns at five unseen incidence angles. Each column corresponds to one incidence angle; the upper and lower rows show the ground-truth pattern $\mat{E}$ and prediction $\widehat{\mat{E}}$.}
\label{fig:incidence_angle_patterns}
\end{figure*}

\begin{table*}[!t]
\caption{Achievable Rate Prediction Across Four Representative RIS-Assisted Communication Scenarios}
\label{tab:communication_rate_results}
\centering
\footnotesize
\setlength{\tabcolsep}{4pt}
\begin{tabular}{lcc>{\bfseries\boldmath}ccc}
\toprule
Scenario & Ground-truth avg. rate & Proposed avg. rate & Proposed error & Ideal-state avg. rate & Ideal-state error \\
 & (Mbit/s) & (Mbit/s) & (\%) & (Mbit/s) & (\%) \\
\midrule
Single-user & 109.677 & 109.691 & 0.012 & 242.929 & 121.494 \\
Dual-user (average) & 54.860 & 55.132 & 0.496 & 149.670 & 172.819 \\
Single-user and Null: intended user & 109.216 & 108.311 & 0.828 & 243.145 & 122.629 \\
Single-user and Null: null direction & 0.081 & 0.073 & 10.110 & 0.873 & 973.579 \\
Six-user (average) & 15.586 & 15.562 & 0.151 & 54.416 & 249.141 \\
\bottomrule
\end{tabular}
\end{table*}

With 200 hardware-impaired training samples, the proposed model outperforms the physics-structured data-only baseline trained on 2000 samples on all three metrics.

\subsection{Incidence Angle Generalization}
To predict RIS responses at unseen incidence angles without additional fine-tuning, we transform the predicted magnetic current to account for changes in the incident spatial phase and polarization.
 The reference incidence angle used for training is
$(\theta_{\mathrm{r}},\varphi_{\mathrm{r}})=(30^{\circ},30^{\circ})$.
Five test datasets are generated at
$(15^{\circ},0^{\circ})$, $(30^{\circ},60^{\circ})$, $(30^{\circ},45^{\circ})$, $(60^{\circ},180^{\circ})$, and $(45^{\circ},30^{\circ})$.
Each dataset contains 100 randomly generated commanded one-bit RIS phase configurations. The same combined impairment sequence and one shared hardware impairment descriptor are used at all five incidence angles, so the evaluation isolates the incidence angle change.

For a given RIS phase configuration, the model predicts the equivalent magnetic current at the reference incidence angle. Then, we adjust this current for the spatial phase and polarization at the target incidence angle and use the adjusted current to calculate the far electric field.
Let $\widehat{\Ms}_{\boldsymbol{\Theta},ij}^{\mathrm{r}}$
denote the predicted magnetic current at the reference
incidence angle. We remove the reference polarization
factor from this current to obtain an effective scalar
coefficient at each RIS element:
\begin{equation}
  \widehat{s}_{ij}
  =
  \frac{
  \widehat{M}_{\boldsymbol{\Theta},x,ij}^{\mathrm{r}}
  E_{0,y}^{\mathrm{r}}
  -
  \widehat{M}_{\boldsymbol{\Theta},y,ij}^{\mathrm{r}}
  E_{0,x}^{\mathrm{r}}
  }
  {
  E_{\mathrm{in}}
  \left[
  (E_{0,y}^{\mathrm{r}})^2
  +
  (E_{0,x}^{\mathrm{r}})^2
  \right]
  },
  \label{eq:angle_s0}
\end{equation}
where $\widehat{M}_{\boldsymbol{\Theta},x,ij}^{\mathrm{r}}$ and $\widehat{M}_{\boldsymbol{\Theta},y,ij}^{\mathrm{r}}$ are the $x$- and $y$-components of the predicted equivalent magnetic current $\widehat{\Ms}_{\boldsymbol{\Theta},ij}^{\mathrm{r}}$. The $E_{0,x}^{\mathrm{r}}$ and $E_{0,y}^{\mathrm{r}}$ are the tangential components of $\vect{E}_{0}^{\mathrm{r}}=[E_{0,x}^{\mathrm{r}},E_{0,y}^{\mathrm{r}},E_{0,z}^{\mathrm{r}}]^T$, which is the unit polarization vector $\vect{E}_0$ in \eqref{eq:einc} evaluated at the reference incidence angle. The incident field magnitude $E_{\mathrm{in}}$ is also defined in \eqref{eq:einc}.
 For a target incidence angle $(\theta',\varphi')$, the change in incident spatial phase relative to the reference incidence angle is defined as
\begin{equation}
  \Delta\psi_{ij}(\theta',\varphi')
  =
  k_0
  \left[
  (k_x'-k_x^{\mathrm{r}})x_{ij}
  +
  (k_y'-k_y^{\mathrm{r}})y_{ij}
  \right],
  \label{eq:angle_phase_delta}
\end{equation}
where $k_0$ is the free-space wavenumber, and $x_{ij}$ and $y_{ij}$
are the coordinates of that element.
The $k_x'=\sin\theta'\cos\varphi'$ and
$k_y'=\sin\theta'\sin\varphi'$ are the $x$- and $y$-components
of the incident unit wave vector at the target incidence
angle, while
$k_x^{\mathrm{r}}=\sin\theta_{\mathrm{r}}\cos\varphi_{\mathrm{r}}$
and
$k_y^{\mathrm{r}}=\sin\theta_{\mathrm{r}}\sin\varphi_{\mathrm{r}}$
are the corresponding components at the reference incidence
angle. 
 The magnetic current used at the queried incidence angle is
\begin{equation}
  \widehat{\Ms}_{\boldsymbol{\Theta},ij}(\theta',\varphi')
  =
  E_{\mathrm{in}}\widehat{s}_{ij}
  \ee^{\jj\Delta\psi_{ij}(\theta',\varphi')}
  [E_{0,y}',-E_{0,x}',0]^T,
  \label{eq:angle_ms_transform}
\end{equation}
where $\vect{E}_{0}'=[E_{0,x}',E_{0,y}',E_{0,z}']^T$ is the unit polarization vector at the queried incidence angle. The transformed magnetic current is then passed through the same fixed Green function propagation operator as in Section~II-B.

Table~\ref{tab:angle_generalization_results} reports the prediction errors at the five tested incidence angles, and Fig.~\ref{fig:incidence_angle_patterns} compares the predicted and ground-truth far electric field magnitude patterns. These results show that the magnetic current transformation maintains prediction accuracy at the tested angles without additional fine-tuning.

\section{Achievable Rate Prediction}

This section evaluates achievable rate prediction for single-user, dual-user, six-user, and single-user-with-null scenarios using the structured beamforming configurations constructed in Appendix~\ref{app:structured_dataset}. The rates at the prescribed receiver directions are calculated from electric fields predicted by the proposed model, which is trained on randomly generated RIS phase configurations and applied without additional fine-tuning.

The electromagnetic setup follows Section~IV-A. The communication parameters are $T=290~\mathrm{K}$, $B=100~\mathrm{MHz}$, and $\mathrm{NF}=7~\mathrm{dB}$. All communication performance evaluations use $G_{\mathrm{f}}[m,n]=1$ and $P_{\mathrm{I}}=0$.

At each receiver location, we compare the achievable rates obtained from the ground-truth field and the fields predicted by the proposed and ideal-state models.
For the single-user scenario, we report the receiver's rate.
For the dual-user and six-user scenarios, we report the
mean rate across the served users.
For the single-user-with-null scenario, we separately report
the rates at the intended receiver and the receiver in the
prescribed null direction.
For each reported rate, the relative prediction error is
$|\widehat{R}-R|/R$, where $R$ is the ground-truth rate
and $\widehat{R}$ is the corresponding predicted rate.

Table~\ref{tab:communication_rate_results} shows that the proposed hardware-aware model closely matches the ground-truth rates across all communication scenarios. In contrast, the ideal-state model consistently overestimates the user rates, highlighting the need for hardware-aware electromagnetic modeling in communication performance prediction.
\section{Conclusion}
This paper developed a physics-informed electromagnetic modeling method for hardware-impaired RISs, combining a two-stage learning strategy with an embedded electromagnetic forward model.
 The method improves prediction accuracy and convergence while enabling far electric field magnitude prediction across different phase configurations without full domain field reconstruction.
  In the evaluated combined impairment setting, the model outperforms the data-only baseline while requiring substantially fewer hardware-impaired training samples.
   The model was also evaluated at unseen incidence angles and on structured beamforming configurations.
   Under the combined impairment setting, its average per-sample inference latency was 5.89~ms.
    Its application to achievable rate prediction without additional fine-tuning demonstrates the value of learning hardware-specific electromagnetic responses for assessing achievable rates and null suppression. These findings support rapid RIS performance evaluation with limited hardware-impaired data and reduced reliance on ideal hardware assumptions.

\appendices
\section{Structured Beamforming Phase Configuration Construction}
\label{app:structured_dataset}
This appendix describes the construction of the structured one-bit RIS phase configurations used in Section~V. We define the spatial phase for a given beam direction, then derive configurations for single-beam, multibeam, and main-lobe-with-null beamforming.

For a beam directed toward $(\theta,\varphi)$, define the spatial phase term at element $(i,j)$ as
\begin{equation}
  \psi(\theta,\varphi;i,j)
  =
  k_0\left(
  x_{ij}\sin\theta\cos\varphi
  +y_{ij}\sin\theta\sin\varphi
  \right),
  \label{eq:app_directional_phase}
\end{equation}
where $\psi(\theta,\varphi;i,j)$ is the spatial phase term
at the $(i,j)$-th RIS element for the beam direction
$(\theta,\varphi)$, $\theta$ and $\varphi$ are the polar
and azimuth angles, $k_0$ is the free-space
wavenumber defined in \eqref{eq:wavenumber}, and $x_{ij}$
and $y_{ij}$ are the coordinates of the element center
along the $x$- and $y$-axes, respectively, as defined in
\eqref{eq:coords}.
 The incident spatial phase is
\begin{equation}
  \psi_{\mathrm{in},ij}
  =
  \psi(\theta_{\mathrm{in}},\varphi_{\mathrm{in}};i,j).
  \label{eq:app_incident_phase}
\end{equation}

For single-beam steering toward $(\theta_1,\varphi_1)$, the continuous phase before one-bit quantization is
\begin{equation}
  \phi_{ij}^{\mathrm{cont}}
  =
  -\psi_{\mathrm{in},ij}
  -\psi(\theta_1,\varphi_1;i,j).
  \label{eq:app_single_beam}
\end{equation}
Equation~\eqref{eq:app_single_beam} compensates the incident phase and aligns the reflected contributions in the target direction.

For multi-beam synthesis with $K_{\mathrm{b}}$ target directions, the continuous complex aperture weight is formed by superposing the steering terms:
\begin{equation}
  \begin{aligned}
  u_{ij}
  &=
  \sum_{h=1}^{K_{\mathrm{b}}}
  w_h\ee^{-\jj\psi(\theta_h,\varphi_h;i,j)},\\
  \phi_{ij}^{\mathrm{cont}}
  &=
  \angle u_{ij}-\psi_{\mathrm{in},ij},
  \end{aligned}
  \label{eq:app_multi_beam}
\end{equation}
where $(\theta_h,\varphi_h)$ is the $h$th beam direction, $w_h$ is its complex aperture weight, and $K_{\mathrm{b}}$ is the number of target beams. The dual-beam and six-beam configurations used in Section~V follow from $K_{\mathrm{b}}=2$ and $K_{\mathrm{b}}=6$, respectively, and generate the phase maps for the dual-user and six-user communication scenarios. 

For beamforming with a main lobe and a prescribed null, the main lobe is directed toward the intended user at $(\theta_{\mathrm{main}},\varphi_{\mathrm{main}})$, and the null is placed at $(\theta_{\mathrm{null}},\varphi_{\mathrm{null}})$. For compactness, write $\psi_{\mathrm{main},ij}=\psi(\theta_{\mathrm{main}},\varphi_{\mathrm{main}};i,j)$ and $\psi_{\mathrm{null},ij}=\psi(\theta_{\mathrm{null}},\varphi_{\mathrm{null}};i,j)$. The main-lobe aperture vector and null steering vector are
\begin{equation}
  \begin{aligned}
  \vect{u}_0
  &=
  \operatorname{vec}\!\left(
  \left\{\ee^{-\jj\psi_{\mathrm{main},ij}}\right\}_{i,j}
  \right),\\
  \vect{a}_{\mathrm{null}}
  &=
  \operatorname{vec}\!\left(
  \left\{\ee^{\jj\psi_{\mathrm{null},ij}}\right\}_{i,j}
  \right),
  \end{aligned}
  \label{eq:app_null_vectors}
\end{equation}
where $\operatorname{vec}(\cdot)$ follows the element order $\ell=(i-1)N+j$ defined in Section~III. To impose a null in the continuous-aperture response, we project $\vect{u}_0$ onto the null space of $\vect{a}_{\mathrm{null}}^{T}$:
\begin{equation}
  \begin{aligned}
  \vect{u}_{\mathrm{proj}}
  &=
  \vect{u}_0-
  \vect{a}_{\mathrm{null}}^{*}
  \frac{\vect{a}_{\mathrm{null}}^{T}\vect{u}_0}
       {\vect{a}_{\mathrm{null}}^{T}\vect{a}_{\mathrm{null}}^{*}},\\
  \phi_{ij}^{\mathrm{cont}}
  &=
  \angle u_{\mathrm{proj},\ell}-\psi_{\mathrm{in},ij}.
  \end{aligned}
  \label{eq:app_null_projection}
\end{equation}
The projection gives $\vect{a}_{\mathrm{null}}^{T}\vect{u}_{\mathrm{proj}}=0$ in the continuous-aperture domain. The continuous phase maps from all four beamforming constructions are converted to commanded one-bit RIS configurations using
\begin{equation}
  Q(\alpha)
  =
  \begin{cases}
    \pi, & 0<\alpha\bmod 2\pi<\pi,\\
    0, & \text{otherwise},
  \end{cases}
  \qquad
  \phi_{ij}=Q\!\left(\phi_{ij}^{\mathrm{cont}}\right).
  \label{eq:app_one_bit_quantizer}
\end{equation}

Finally, the commanded one-bit phase configurations are passed through the combined impairment model in Section~III-F to generate the corresponding far electric field magnitude responses.

\end{document}